\documentclass[journal]{IEEEtran}
\usepackage{amsmath,amsfonts}
\usepackage{algorithmic}
\usepackage{algorithm}
\usepackage{graphicx}
\usepackage{listings} 
\usepackage[table,xcdraw]{xcolor}
\usepackage[table,xcdraw]{xcolor} 
\usepackage{colortbl} 
\usepackage[compatibility=false]{caption}
\usepackage{multirow}
\usepackage{graphicx}
\usepackage{array}
\usepackage{textcomp}
\usepackage{float}
\usepackage{subcaption}
\usepackage{color}
\usepackage{url}
\usepackage{multirow}
\usepackage{verbatim}
\usepackage{graphicx}
\usepackage{cite}
\usepackage{colortbl}
\usepackage{booktabs}
\definecolor{lightblue}{RGB}{230, 240, 255}
\definecolor{lightyellow}{RGB}{255, 250, 230}
\definecolor{lightgray}{RGB}{245, 245, 245}
\begin{document}

\title{RefactorCoderQA: Benchmarking LLMs for Multi-Domain Coding Question Solutions in Cloud and Edge Deployment}

\author{Shadikur Rahman, Aroosa Hameed, Gautam Srivastava, Syed Muhammad Danish 
\IEEEcompsocitemizethanks{
\IEEEcompsocthanksitem Shadikur Rahman is with York University and Algoma University, Toronto, Canada. (email: shadikur@yorku.ca, shadikur.rahman@algomau.ca)
\IEEEcompsocthanksitem Syed Muhammad Danish is with Algoma University, Brampton, Canada. (email: syed.danish@algomau.ca)
\IEEEcompsocthanksitem Aroosa Hameed is with Department of Systems and Computer Engineering, Carleton University, Ottawa, Canada. (email: aroosahameed@cunet.carleton.ca).
\IEEEcompsocthanksitem Gautam Srivastava is with Department of Math and Computer Science, Brandon University, Brandon, Canada (email: srivastavag@brandonu.ca).
}}



\maketitle

\begin{abstract}

To optimize the reasoning and problem-solving capabilities of Large Language Models (LLMs), we propose a novel cloud–edge collaborative architecture that enables a structured, multi-agent prompting framework. This framework comprises three specialized components: GuideLLM, a lightweight model deployed at the edge to provide methodological guidance; SolverLLM, a more powerful model hosted in the cloud responsible for generating code solutions; and JudgeLLM, an automated evaluator for assessing solution correctness and quality. To evaluate and demonstrate the effectiveness of this architecture in realistic settings, we introduce RefactorCoderQA, a comprehensive benchmark designed to evaluate and enhance the performance of LLMs across multi-domain coding tasks. Motivated by the limitations of existing benchmarks, RefactorCoderQA systematically covers various technical domains, including Software Engineering, Data Science, Machine Learning, and Natural Language Processing, using authentic coding challenges from Stack Overflow. We propose \textit{RefactorCoder-MoE}, a fine-tuned mixture-of-experts (MoE) code language model based on DeepSeek-Coder-7B-Instruct, adapted to the RefactorCoderQA benchmark using QLoRA for domain-specific coding question answering. Extensive experiments reveal that RefactorCoder-MoE achieves strong and competitive performance, significantly outperforming all evaluated open-source and commercial baselines, with an overall accuracy of 76.84\%. 

\end{abstract}


\section{Introduction}

In recent years, Large Language Models (LLMs) have become valuable tools for companies aiming to speed up software development, improve data analysis, and expand the use of artificial intelligence. These models are now commonly used in business settings to perform tasks that once required experienced developers or data scientists \cite{chowdhary2025autopac}. Examples include writing and fixing code, analyzing data, creating technical documentation, and improving machine learning models. Because LLMs can understand both natural language and programming code, they help teams work more efficiently, release products faster, and support collaboration across different roles \cite{schnepf2024studies}.

As companies continue to rely on LLMs for development tasks, either through cloud-based APIs or by deploying lightweight models on edge devices, one of the main difficulties they face is evaluating how effectively these models handle real-world, complex coding problems \cite{fodor2025line}. Although LLMs have made great progress, most existing benchmarks are still limited. They often focus on just one programming language, involve only simple or standalone problems, and do not follow a clear method for evaluating results. Because of this, it is difficult to tell whether a model can perform tasks that professionals face in practice, such as using different tools, understanding the context of the code, or solving problems that require several steps. To better test LLMs in realistic situations, there is a need for richer and more diverse benchmarks that cover multiple technical areas and a wider range of task types \cite{austin2021program, li2022competition}. This leads to out first research question: how can we build a benchmark that reflects real-world, multi-domain coding problems and provides a fair and useful way to measure LLM performance?

\begin{figure}[t!]
\centering
\includegraphics[width=\linewidth]{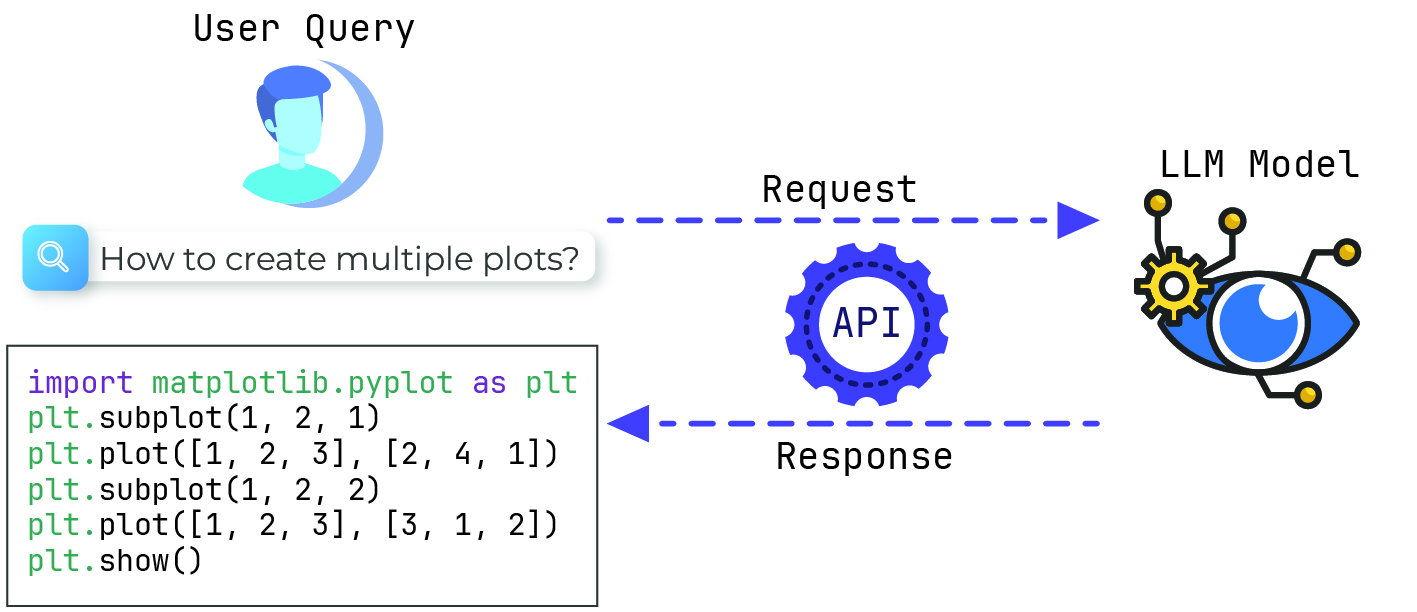}
\caption{
LLM raw response to the query
}
\label{fig:raw-framework}
\end{figure}

In addition to the limits of current benchmarks, the way tasks are given to LLMs also has a big impact on how well they perform. In real-world settings, when models receive raw or poorly structured prompts, they often return incomplete or overly simple answers. For example, as illustrated in Fig. \ref{fig:raw-framework}, when LLMs are given unstructured prompts, they often produce low-quality responses that are difficult to analyze or compare. A typical case is when a model generates a basic plot but leaves out important details like axis labels or proper formatting, making the output unclear and less useful. Such incomplete responses only cover part of the task and fall short in situations where careful reasoning and attention to detail are essential.  This shows that we need better ways to test LLMs and guide them through structured prompts that reflect the real challenges developers face in both centralized and edge-based development workflows. This leads to our second research question: How can we design prompting and evaluation methods that help LLMs produce complete, accurate, and useful solutions for real-world industry tasks?

To address the first research question, we introduce RefactorCoderQA, a comprehensive and diverse benchmark designed to evaluate and enhance the performance of LLMs on coding tasks. This benchmark covers four key technical domains: Software Engineering (SE), Data Science (DS), Machine Learning (ML), and Natural Language Processing (NLP). RefactorCoderQA is built using 2,635 real-world coding questions sourced from Stack Overflow, a widely-used platform known for its practical, community-driven discussions \cite{ragkhitwetsagul2019toxic, zhang2019empirical}. Each question includes a detailed problem statement along with an accepted solution, and has been reformatted into a consistent input-output format to support structured prompting and objective evaluation. The dataset was developed through a systematic process of data collection, cleaning, and organization. By using real questions and answers from actual development scenarios, RefactorCoderQA offers a more realistic and meaningful way to assess the capabilities of LLMs across a wide range of domains and coding challenges.

\begin{figure*}[t!]
\centering
\includegraphics[width=\linewidth]{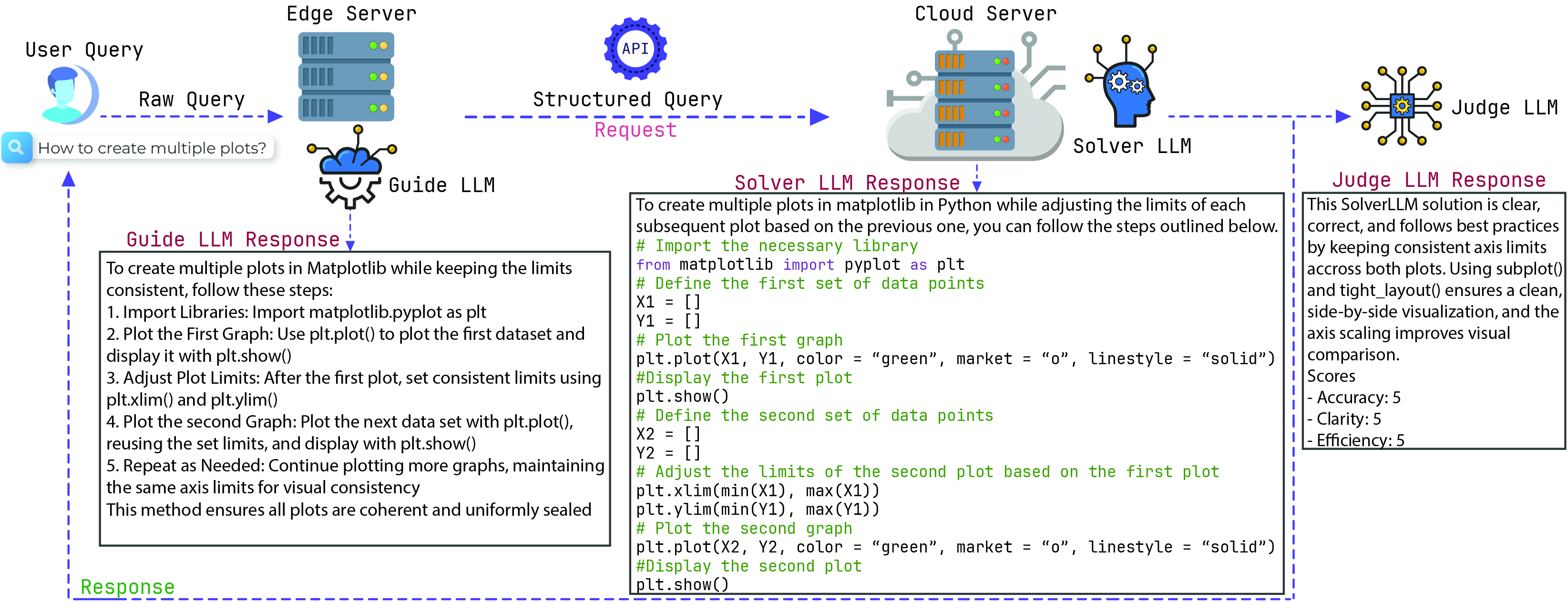}
\caption{
Overview of the RefactorCoderQA agentic framework. The process begins with a problem statement and flows through three stages: \textbf{GuideLLM} (methodology generation), \textbf{SolverLLM} (solution synthesis), and \textbf{JudgeLLM} (automated evaluation).
}
\label{fig:agentic-framework}
\end{figure*}

To address the second research question, we propose a novel
cloud–edge collaborative architecture that enables a structured,
multi-agent prompting framework. We adopted structured prompting, a method that has been shown to significantly improve LLM performance and interpretability, especially for complex, domain-specific coding tasks \cite{brown2020language, wei2022chain, wang2022self}. As illustrated in Fig. \ref{fig:agentic-framework}, our proposed multi-agent framework, integrated into the RefactorCoderQA pipeline, decomposes the code-generation process into three specialized LLM roles to improve robustness and solution quality. The \textit{GuideLLM}, powered by the fine-tuned RefactorCoder-MoE model, processes raw user queries and converts them into structured, well-defined requests by identifying the user intent, relevant constraints, and expected output format. This structured guidance enables the \textit{SolverLLM}, which also leverages RefactorCoder-MoE, to focus on generating accurate and executable code rather than interpreting ambiguous natural-language queries. Finally, the \textit{JudgeLLM}, built on GPT-4o, evaluates the generated solutions in terms of correctness, clarity, and efficiency, and provides detailed feedback. This design addresses the limitations of single, general prompts, which often lead to incomplete or poorly structured outputs. Decomposing the task into guiding, solving, and evaluation stages enables clearer reasoning, higher-quality code generation, and more reliable assessment.
To summarize, this paper makes the following contributions:
\begin{itemize}
    \item We introduce \textit{RefactorCoderQA}, a benchmark dataset built from real-world coding questions, designed to evaluate LLMs across multiple technical domains, including software engineering, data science, machine learning, and natural language processing, in realistic scenarios. We make this dataset openly available at \footnote{\url{https://github.com/sadirahman/RefactorCoderQA}}.

    \item We propose a structured three-part prompting framework that guides LLMs through coding tasks using explicit planning, execution, and evaluation stages, improving solution quality, interpretability, and evaluation consistency across diverse coding scenarios. Through comprehensive empirical evaluation, we demonstrate the effectiveness of both the proposed framework and RefactorCoder-MoE, showing strong and consistent improvements over all evaluated baselines across domains and task types.
\end{itemize}

The remainder of the paper is organized as follows. Section II reviews related work. Section III describes the proposed methodology behind \textit{RefactorCoderQA}. Section IV presents the RefactorCoder framework. Section V presents the implementation details and experimental results. Section VI outlines the limitations of the study. Finally, Section VII concludes the paper and highlights directions for future work.

\section{Related Work}

As LLMs continue to improve, there is a growing need for reliable benchmarks to evaluate their performance in code understanding and generation. While several datasets have been introduced for this purpose, most are limited to specific domains or individual tasks and do not capture the complexity or variety of real-world programming challenges.

The CS1QA benchmark, introduced in \cite{lee2022cs1qa}, focuses on evaluating code-based question answering within educational settings. It includes a collection of 9,237 annotated question-answer pairs collected from beginner-level Python programming courses. The benchmark tests models on various tasks, including classifying question types, selecting relevant lines of code, and retrieving appropriate answers. These tasks highlight the challenge of combining natural language processing with code understanding. While useful for measuring foundational skills, its focus on educational content limits its applicability to broader, real-world settings. Another notable benchmark is the SWE-bench \cite{jimenez2023swe}, which targets real-world scenarios involving code generation and bug fixing by using publicly accessible pull requests from well-known Python projects to perform evaluations. Although the benchmark provides a thorough testing ground, its effectiveness is somewhat limited by its dependence on open source contributions and the restricted context-handling abilities of current language models. CodeQueries \cite{sahu2024codequeries} takes a semantic approach by testing LLMs on their ability to perform extractive question answering over Python code. With 52 queries requiring both single-hop and multi-hop reasoning, it focuses on precise code comprehension but remains constrained to a single language and query type.

CoReQA \cite{chen2025coreqa} presents a valuable benchmark for assessing language models on repository-level question answering using real GitHub issues and comments. It captures the complexity of practical software development queries across multiple programming languages and domains. The benchmark also introduces a robust evaluation framework using LLM-as-a-judge to assess answer quality. However, results show that even advanced models struggle without effective context retrieval, revealing a significant limitation in current LLM capabilities. In the area of program repair, MOREPAIR \cite{yang2024multi} introduces a multi-objective fine-tuning framework that leverages natural language explanations to guide code fixes, making it adaptable across programming languages. By combining code transformation learning with conversational guidance, it achieves strong performance even with smaller datasets. However, its effectiveness still depends on the quality and relevance of the generated natural language guidance.

Lajkó et al. \cite{lajko2022fine} enhanced GPT-2 by fine-tuning it on 16,000 JavaScript code samples. They assessed both the original and fine-tuned versions using a dataset of 18,736 examples derived from GitHub commits, allocating 16,863 samples for training and 1,559 for testing. The fine-tuned model achieved a 15.5\% gain in TOP-10 accuracy on a JavaScript-specific benchmark. Jiang et al. \cite{jiang2023impact} investigated how large language models influence automated program repair (APR) and assessed the effects of fine-tuning these models across four APR benchmarks. To address concerns related to data leakage, they also introduced a new benchmark, HumanEval-Java, for more reliable evaluation. RepairLLaMA \cite{silva2023repairllama} proposed an innovative fine-tuning method for automated program repair that integrates tailored code representations with optimized fine-tuning strategies. This method enabled the model to better align large language models with the requirements of code repair, leading to substantial performance improvements over the CodeLlama-13B baseline on several Java benchmarks.

DebugBench \cite{tian2024debugbench} is a comprehensive dataset consisting of 4,253 code snippets designed to evaluate LLMs on diverse debugging tasks. It features bug-injected solutions across multiple languages and follows a detailed taxonomy to categorize bug types. A limitation is that the reliance on GPT-4 for bug generation may introduce patterns that differ from real-world coding errors. Lastly, InfiBench \cite{li2024infibench} introduces a comprehensive and realistic benchmark for evaluating code LLMs through free-form question answering, capturing how developers interact with models in real-world settings. It spans multiple languages and domains, using diverse evaluation metrics beyond traditional code generation tasks. However, its evaluation still faces challenges in defining universal correctness for open-ended responses, which can impact consistency across different tasks.

\subsection{Novelty}
Although existing benchmarks have substantially advanced the evaluation of LLMs on individual coding tasks such as educational QA (CS1QA), program repair (MOREPAIR, RepairLLaMA), semantic code querying (CodeQueries), and repository-level reasoning (CoReQA), they are typically constrained in scope. Many are limited to a single programming language (e.g., Python or JavaScript), focus on isolated capabilities (e.g., bug fixing or extractive QA), or operate within tightly bounded domains such as education or open-source projects. While benchmarks like SWE-bench and InfiBench attempt to bridge the gap between academic settings and real-world development, they still struggle to represent full range of developer activities and complete development workflows.

Recent agentic prompting approaches, such as Chain-of-Thought (CoT)~\cite{wei2022chain}, Self-Consistency~\cite{chen2023universal}, Plan-Execute frameworks (e.g., ReAct\cite{yao2022react}), and Reflexion~\cite{shinn2023reflexion}, improve reasoning by encouraging a single model to generate intermediate reasoning steps, decompose tasks into planning and execution phases, or iteratively refine outputs through self-feedback. While effective for enhancing reasoning consistency, these methods typically rely on a single model and do not explicitly separate planning, execution, and evaluation into specialized roles.


\begin{table*}[t]
\centering
\caption{Comparison with prior agentic prompting methods}
\label{tab:agentic-comparison}
\renewcommand{\arraystretch}{1.1}
\begin{tabular}{>{\raggedright\arraybackslash}p{3.4cm}cccccc}
\toprule
\textbf{Method} & 
\textbf{Role Separation} & 
\textbf{Planning} & 
\textbf{Execution} & 
\textbf{Independent Evaluation} &
\textbf{Domain-Specific Guidance}  \\
\midrule
Chain-of-Thought~\cite{wei2022chain} & 
\texttimes & \checkmark & \checkmark & \texttimes & \texttimes  \\

Self-Consistency~\cite{chen2023universal} & 
\texttimes & \checkmark & \checkmark & \texttimes & \texttimes   \\

Plan-Execute~\cite{yao2022react} & 
\texttimes & \checkmark & \checkmark & \texttimes & \texttimes   \\

Reflexion~\cite{shinn2023reflexion} & 
\texttimes & \checkmark & \checkmark & \texttimes & \texttimes   \\

RefactorCoder (Ours) & 
\checkmark & \checkmark & \checkmark & \checkmark & \checkmark  \\
\bottomrule
\end{tabular}
\end{table*}

In contrast, RefactorCoderQA adopts an explicit multi-agent design in which \textit{GuideLLM} generates structured methodological guidance, \textit{SolverLLM} executes the solution conditioned on this guidance, and \textit{JudgeLLM} provides an independent evaluation signal. This explicit role separation enables modular reasoning, reduces error propagation, and supports reproducible evaluation, distinguishing our framework from prior single-model or self-consistency-based prompting strategies. A summary comparison with representative agentic prompting methods is provided in Table~\ref{tab:agentic-comparison}.


\section{RefactorCoderQA Benchmark}
We present the methodology behind the creation of RefactorCoderQA,  followed by a comprehensive description of our data collection pipeline.

\begin{table}[ht!]
\centering
\small
\caption{RefactorCoderQA Dataset Summary by Domain and Topic (Filtered)}
\label{tab:dataset-summary}
\resizebox{\columnwidth}{!}{%
\begin{tabular}{llccc}
\toprule
\textbf{Domain} & \textbf{Topic} & \textbf{Collected} & \textbf{Filtered} & \textbf{Domain Total} \\
\midrule
\multirow{4}{*}{SE} 
& Python & 203 & 125 & \multirow{4}{*}{677} \\
& C & 296 & 234 & \\
& Java & 257 & 172 & \\
& JavaScript & 224 & 146 & \\
\midrule
\multirow{4}{*}{DS} 
& NumPy & 346 & 231 & \multirow{4}{*}{827} \\
& Pandas & 338 & 219 & \\
& Matplotlib & 296 & 212 & \\
& Data Science & 290 & 165 & \\
\midrule
\multirow{5}{*}{ML} 
& TensorFlow & 192 & 99 & \multirow{5}{*}{626} \\
& PyTorch & 195 & 105 & \\
& Scikit-learn & 281 & 193 & \\
& Cross-validation & 267 & 184 & \\
& Fine-tuning & 76 & 45 & \\
\midrule
\multirow{6}{*}{NLP} 
& LLM & 116 & 54 & \multirow{6}{*}{505} \\
& Llama & 112 & 58 & \\
& Hugging Face & 170 & 85 & \\
& Multimodal & 12 & 9 & \\
& ChatGPT & 174 & 78 & \\
& Prompt Eng. & 266 & 173 & \\
\midrule
\multicolumn{2}{l}{\textbf{Overall Total}} & \textbf{4352} & \textbf{2635} & \textbf{2635} \\
\bottomrule
\end{tabular}
}
\end{table}


\subsection{Criteria for RefactorCoderQA Tasks}
Our focus is on developing realistic, diverse, and practically relevant coding-related tasks suitable for effectively evaluating and refining LLMs. To ensure the quality and utility of tasks included in RefactorCoderQA, we established the following explicit criteria:

\begin{itemize}
    \item \textit{Realistic}: Tasks reflect authentic coding challenges encountered by software engineers, data scientists, and machine learning practitioners, ensuring practical relevance.
    
    \item \textit{Diverse Domains}: Tasks cover multiple critical domains such as DS, SE, ML, and NLP, ensuring comprehensive evaluation across relevant areas.
    
    \item \textit{Clearly Structured}: Each task is systematically organized into an explicit title-description-acceptable answer format, facilitating structured evaluation and prompt-based interactions.
    
    \item \textit{Automatically Evaluation}: Tasks have clear solutions that can be automatically and objectively evaluated for correctness, completeness, and accuracy, enabling reliable benchmarking.
\end{itemize}
This structure ensures the robustness and practical applicability of the RefactorCoderQA dataset, providing a strong foundation for assessing and advancing LLM capabilities.

\subsection{Data Collection}

To ensure diversity and practical relevance, RefactorCoderQA includes carefully curated coding questions from Stack Overflow across four key domains: Software Engineering, Data Science, NLP, and ML.

\begin{figure*}[t]
\centering
\includegraphics[width=1.0\linewidth]{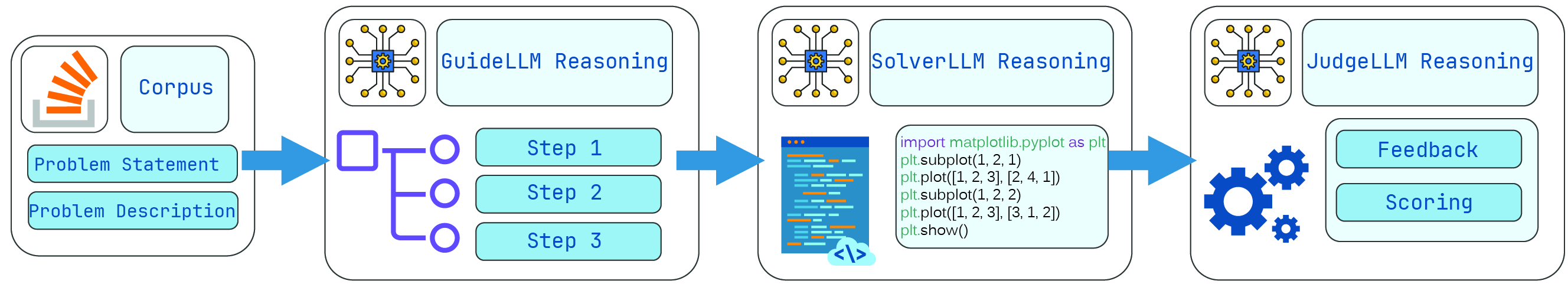}
\caption{Overview of the RefactorCoder Agentic Framework Workflow: A multi-agent framework that processes coding-related problem statements through three coordinated stages, GuideLLM for structured methodology generation, SolverLLM for executable code synthesis, and JudgeLLM for automated evaluation across accuracy, clarity, and efficiency dimensions.}
\label{fig:code-generation}
\end{figure*}

\textit{SEQA:} We collected SE questions from Stack Overflow, focusing on practical tasks including debugging, algorithm optimization, coding solutions, and architecture-related issues. Specifically, our dataset includes questions on programming languages like Python (203), C (296), Java (257), and JavaScript (224), totalling \textbf{980 questions}. 

\textit{DSQA:} In the DS domain, we created questions primarily involving data manipulation, pre-processing, analysis, and visualization. The data collected consists of questions on widely used libraries such as NumPy (346), Pandas (338), Data Science (290) and Matplotlib (296). Overall, the DS portion comprises \textbf{1270 questions}.

\textit{NLPQA:} For the NLP section, questions were collected focusing on language processing tasks, including text preprocessing, language modelling, prompt engineering, and libraries or frameworks commonly used in NLP workflows. Specific topics collected include Hugging Face (170), ChatGPT (174), Prompt Engineering (266), Multimodal (12), Llama (112), and LLM (116), totaling \textbf{ 850 questions}. 

\textit{MLQA:} Questions related to ML covered essential aspects of the ML pipeline, such as model building, training, fine-tuning, validation, and deployment. The data set includes tasks on popular ML frameworks such as TensorFlow (192), PyTorch (195), Scikit-learn (281), Fine-tuning(76) and Cross-validation (267). In total, the Machine Learning dataset consists of \textbf{1011 questions}.

\subsection{Data Analysis}\label{data analysis}

To ensure the quality, reliability, and reproducibility of RefactorCoderQA, we performed a comprehensive analysis and refinement of the initially collected dataset. Our analysis involved multiple stages, from automated filtering to manual validation, resulting in a carefully curated benchmark. 

First, we relied on Stack Overflow’s \textit{accepted answer} information, which is directly available through the Stack Overflow API. During data extraction, we selected only questions that included an accepted answer, ensuring that each entry had a solution explicitly confirmed by the original poster as resolving the issue. This step guaranteed that all problem statements in the dataset were realistic and demonstrably answerable.

Next, we performed a detailed \textbf{manual analysis} of the collected entries. Questions that lacked sufficient context, were ambiguous, or included incomplete solutions were removed. We prioritized problem statements with clear descriptions and answers that demonstrated step-by-step, implementable resolutions. This process strengthened the dataset’s usability for real-world programming tasks.

To further enhance consistency, we applied a structured data cleaning and normalization pipeline. Duplicate rows were eliminated, special characters were stripped, and titles were normalized to lowercase with uniform formatting. Each record was reformatted into a triplet structure: 
\textbf{Instruction} (cleaned question title), 
\textbf{Input} (detailed problem description), and 
\textbf{Output} (the accepted answer). 
This structured representation facilitates fine-tuning for large language models and ensures the dataset’s readiness for downstream evaluation.

Finally, we conducted a manual categorization process. Each problem statement was labelled according to both domain-level categories (Software Engineering, Data Science, Machine Learning, Natural Language Processing) and topic-level categories (e.g., NumPy, Hugging Face, TensorFlow). This step, verified by multiple annotators, ensured balanced representation across domains and improved the interpretability of evaluation results. 

Table~\ref{tab:dataset-summary} provides a summary of the dataset before and after filtering. The reduction from 4,352 initially collected questions to 2,635 high-quality entries demonstrates the thoroughness of our analysis, with only well-defined and answerable problems retained for RefactorCoderQA.

\section{RefactorCoder Framework}

RefactorCoder code generation pipeline uses a structured multi-agent framework powered by specialized language models, especially our fine-tuned model, \textbf{RefactorCoder}, to guide and evaluate the creation of coding solutions.

As illustrated in Fig. \ref{fig:code-generation}, the process begins with input queries comprising problem statements and descriptions derived from real-world developer discussions. These inputs are processed by the GuideLLM agent, built on RefactorCoder-MoE, which produces structured, multi-step reasoning instructions to guide the subsequent code generation process. The \textbf{SolverLLM} component, also based on RefactorCoder-MoE, then synthesizes executable code by following these instructions generated by the GuideLLM. Finally, the \textbf{JudgeLLM} module, utilizing GPT-4o, performs automated evaluation of the generated code produced by the SolverLLM, assessing it for correctness, clarity, and efficiency. This three-stage design simulates real-world developer workflows while enabling rigorous assessment of LLM capabilities in practical coding tasks.

RefactorCoderQA covers four major domains central to software development and AI applications: SE, DS, ML, and NLP. For each domain, we collected realistic, developer-facing tasks from Stack Overflow, ensuring high relevance and diversity.

\begin{itemize}
    \item \textit{SE tasks} include coding tasks related to algorithm implementation (e.g., reversing linked lists in Java), debugging (e.g., segmentation faults in C), refactoring (e.g., improving Python function readability), and architectural decision-making.
    
    \item \textit{DS tasks} include coding tasks related to merging and cleaning Pandas DataFrames, fixing NumPy broadcasting issues, and customizing Matplotlib plots for clarity.

    \item \textit{ML tasks} cover coding tasks related to custom loss function design in TensorFlow, resolving training errors in PyTorch, and performing cross-validation with hyperparameter tuning in Scikit-learn.

    \item \textit{NLP tasks} involve coding tasks related to prompt engineering, transformer model usage (e.g., Hugging Face, LangChain), and fine-tuning LLMs like Llama for specialized NLP use cases.
\end{itemize}

To ensure a robust and interpretable assessment of LLM reasoning across all domains, our framework evaluates model outputs along two key dimensions:
\begin{table}[t!]
\centering
\small
\caption{Dataset Split Summary for RefactorCoder-MoE Fine-tuning}
\label{tab:data-split-summary}
\resizebox{\columnwidth}{!}{%
\begin{tabular}{lccccc}
\toprule
\textbf{Split} & \textbf{SEQA} & \textbf{NLPQA} & \textbf{MLQA} & \textbf{DSQA} & \textbf{OverallQA} \\
\midrule
Training   & 544 & 416 & 501 & 668 & 2090 \\
Validation & 69  & 52  & 63  & 84  & 262 \\
Test       & 68  & 52  & 63  & 84  & 261 \\
\midrule
\textbf{Total} & \textbf{681} & \textbf{520} & \textbf{627} & \textbf{836} & \textbf{2613} \\
\bottomrule
\end{tabular}
}
\end{table}

\subsubsection*{1. Solution Correctness and Execution Validation}
We define domain-specific evaluation criteria such as successful code execution, correctness against benchmarks or unit tests, runtime performance, and adherence to best practices. Each solution is assessed using automated Python scripts to ensure only fully executable and standards-compliant outputs are considered correct.

\subsubsection*{2. Structured Reasoning Formatting}
Our proposed model, \textbf{RefactorCoder}, consists of three-stage multi-agent reasoning pipeline: the \textit{GuideLLM} provides explicit methodological guidance, the \textit{SolverLLM} generates optimized code, and the \textit{JudgeLLM} evaluates outputs against structured criteria. Automated scripts strictly enforce this reasoning format, any deviation, even if the solution is functionally correct, is marked incorrect. This design supports greater interpretability and ensures consistent and transparent evaluation across all domains.

\subsection{Agent Architecture and Implementation Details}

The RefactorCoder framework adopts a modular multi-agent architecture composed of three components: \textit{GuideLLM}, \textit{SolverLLM}, and \textit{JudgeLLM}, which operate sequentially within a unified inference pipeline, as illustrated in Fig.~\ref{fig:code-generation}. This explicit role separation enables structured planning, controlled solution generation, and independent evaluation.

\textbf{GuideLLM} is responsible for high-level problem understanding and methodological planning. Given the input query, it uses the fine-tuned RefactorCoder-MoE model to generate structured guidance via prompt conditioning, which is then used to condition SolverLLM during solution generation. \textbf{SolverLLM} performs solution execution by generating the final response conditioned on both the original query and the guidance produced by GuideLLM. It uses the fine-tuned RefactorCoder-MoE model trained with QLoRA on the RefactorCoderQA dataset. \textbf{JudgeLLM} provides independent evaluation of SolverLLM outputs and is implemented using GPT-4o. It scores responses along correctness, clarity, and efficiency and is used exclusively for benchmarking, without participating in training or inference-time generation.

At inference time, each query passes sequentially through GuideLLM, SolverLLM, and JudgeLLM. This modular design improves interpretability and reproducibility while allowing individual agents to be modified or replaced independently.

\subsection{RefactorCoder MOE Fine Tune Model}\label{RefactorCoderLLm}
The primary objective of fine-tuning \textbf{RefactorCoder-MoE} is to significantly enhance its performance and specialization in solving realistic coding problems across multiple domains by leveraging domain-specific knowledge extracted from real-world questions and solutions, primarily sourced from Stack Overflow. By adapting the RefactorCoder-MoE model to specific contexts and structured solution guidelines, the fine-tuning process aims to improve the model's capability to understand complex instructions, reason effectively, generate accurate code implementations, and deliver practical, interpretable solutions aligned with professional coding standards.

\subsubsection{Data Split}

We partition the RefactorCoderQA dataset into training, validation, and testing sets using an 80\%-10\%-10\% split across four domains: SE, DS, NLP, and ML. Additionally, an \textbf{OverallQA} set aggregates all domains to support comprehensive evaluation. The training set is used to optimize model parameters, the validation set assists in hyperparameter tuning and early stopping, and the test set offers an unbiased estimate of generalization performance. This consistent partitioning strategy ensures fair comparisons across domains while maintaining domain diversity within each split. Table~\ref{tab:data-split-summary} summarizes the distribution of problem statements across the different domains.

\begin{table}[h!]
\centering
\small
\caption{Fine-Tuning Configuration for RefactorCoder-MoE}
\label{tab:finetuning-config}
\begin{tabular}{ll}
\toprule
\textbf{Parameter} & \textbf{Value} \\ 
\midrule
Base Model & \texttt{deepseek-coder-7b} \\
Input Format & JSONL (Instruction--Input--Output) \\
Max Sequence Length & 512 tokens \\
Batch Size (per device) & 2 \\
Gradient Accumulation Steps & 4 \\
Effective Batch Size & 8 \\
Epochs & 5 \\
Precision & Mixed (fp16) \\
Optimizer & AdamW \\
Learning Rate & $2\times10^{-5}$ \\
Weight Decay & 0.01 \\
LR Scheduler & Cosine (no warmup) \\
LoRA Rank ($r$) & 8 \\
LoRA Scaling Factor ($\alpha$) & 16 \\
LoRA Dropout & 0.1 \\
Hardware & 4 $\times$ NVIDIA A100 GPUs (40GB) \\
Early Stopping & Patience = 2 epochs \\
Checkpointing & Best model retained, others pruned \\
\bottomrule
\end{tabular}
\end{table}

\subsubsection{Fine-Tuning Process}

We fine-tuned \textbf{RefactorCoder-MoE} using the \texttt{deepseek-coder-7b} checkpoint~\cite{guo2024deepseek} as the base model. The training corpus comprised \textbf{2,635 question–answer pairs} from the RefactorCoderQA dataset, curated through the data collection and analysis pipeline described in Section~\ref{data analysis}. Each sample was structured into a JSONL format with three standardized fields: \texttt{Instruction} (problem title), \texttt{Input} (detailed description), and \texttt{Output} (accepted solution). The inputs were tokenized with a maximum sequence length of $512$ tokens. We summarize the key conclusions of the token-length analysis here and defer detailed statistics to Appendix~B (Table~\ref{tab:token-length-stats}). An empirical analysis of GuideLLM and SolverLLM outputs shows that over $95\%$ of responses fall within this limit, with mean lengths below $300$ tokens across all domains. This choice therefore, sufficiently covers both planning and solution-generation outputs while maintaining computational efficiency and consistent evaluation. Dataset partitioning followed the $80/10/10$ split shown in Table~\ref{tab:data-split-summary}, enabling consistent evaluation across the SE, DS, NLP, and ML domains.

\textbf{Model Configuration:}  
We employed 4-bit QLoRA~\cite{dettmers2023qlora} with the PEFT library to achieve parameter-efficient fine-tuning. LoRA adapters were configured with rank $r=8$, scaling factor $\alpha=16$, and dropout $0.1$, while the base model parameters remained frozen. This design reduced memory usage without compromising performance. Tokenization and training were implemented using the Hugging Face \texttt{transformers} and \texttt{datasets} libraries.

\textbf{Training Setup and Optimization:}  
Fine-tuning was carried out on 4 NVIDIA A100 GPUs (40GB VRAM each) using mixed-precision (fp16). An effective batch size of $8$ was achieved via a per-device batch size of $2$ with gradient accumulation over $4$ steps. Training was run for $5$ epochs, with evaluation and checkpointing at the end of each epoch.We used the AdamW optimizer with a $2\times10^{-5}$ learning rate, $0.01$ weight decay, and cosine scheduling (no warmup).Early stopping (patience $=2$) and checkpoint pruning were applied to avoid overfitting and minimize storage.

This end-to-end configuration ensured a reproducible fine-tuning pipeline that balanced computational efficiency with model accuracy, thereby enhancing RefactorCoder-MoE’s ability to solve realistic coding problems across multiple domains.





\bigskip
\fbox{%
\begin{minipage}{0.93\columnwidth}
\textbf{Example 1: GuideLLM Reasoning Prompt}

\smallskip
\textbf{Problem Statement:} how to create multiple plots?

\smallskip
\textbf{Description:} I'm to Python and learning it by doing. I want to make two plots with matplotlib in Python. The second plot keeps the limits of first one. Wonder how I can change the limits of each next plot from previous. Any help, please. What is the recommended method? 

X1 = [80, 100, 120, 140, 160, 180, 200, 220, 240, 260]

Y1 = [70, 65, 90, 95, 110, 115, 120, 140, 155, 150]

from matplotlib import pyplot as plt

plt.plot(
    X1
  , Y1
  , color = "green"
  , marker = "o"
  , linestyle = "solid"
)
plt.show()

X2 = [80, 100, 120, 140, 160, 180, 200]
Y2 = [70, 65, 90, 95, 110, 115, 120]

plt.plot(
    X2
  , Y2
  , color = "green"
  , marker = "o"
  , linestyle = "solid"
)
plt.show()

\smallskip
\textbf{Instruction to Model:}  
\textit{Provide a detailed, step-by-step methodological guide outlining how to approach and effectively solve the problem described above. Clearly identify key steps, logical processes, essential considerations, and conceptual frameworks necessary to address the task comprehensively. Avoid providing the final code implementation, focusing instead on a structured, clear, and actionable methodological outline.}
\end{minipage}
}
\bigskip
\subsection{GuideLLM Reasoning}\label{Guidellm}


GuideLLM serves as the first stage in our multi-agent prompting framework, generating structured, high-level reasoning before any code is produced. Rather than outputting executable code, it focuses on understanding the problem, decomposing it into smaller steps, and outlining a clear solution plan. This reasoning process is handled by our fine-tuned \textit{RefactorCoder-MoE} model, which is trained to provide interpretable and domain-relevant guidance across diverse coding tasks. 

Each input to GuideLLM follows a standardized format that includes a problem statement, a description of the task, and an instruction to generate a step-by-step solution plan. As illustrated in \textbf{Example 1}, the model is explicitly instructed to avoid code generation and instead return a structured methodology. This design encourages conceptual clarity, careful reasoning, problem decomposition,  and essential elements for reliable downstream code generation. By decoupling task interpretation from code synthesis, GuideLLM improves reasoning transparency, reduces hallucinations, and enables more faithful alignment between problem requirements and generated solutions. The resulting guidance is then passed to SolverLLM, which generates code aligned with the outlined steps.

\subsection{SolverLLM Reasoning for Code Generation} 

SolverLLM is responsible for generating executable code solutions based on the structured reasoning produced by GuideLLM. This component is implemented using our fine-tuned \textit{RefactorCoder-MoE} model, trained to follow task-specific instructions and generate high-quality code.

\bigskip
\fbox{%
\begin{minipage}{0.93\columnwidth}
\textbf{Example 2: SolverLLM Reasoning Prompt}

\smallskip
\textbf{Problem Statement:} how to create multiple plots?

\smallskip
\textbf{Description:} I'm to Python and learning it by doing. I want to make two plots with matplotlib in Python. The second plot keeps the limits of first one. Wonder how I can change the limits of each next plot from previous. Any help, please. What is the recommended method? 

X1 = [80, 100, 120, 140, 160, 180, 200, 220, 240, 260]

Y1 = [70, 65, 90, 95, 110, 115, 120, 140, 155, 150]

from matplotlib import pyplot as plt

plt.plot(
    X1
  , Y1
  , color = "green"
  , marker = "o"
  , linestyle = "solid"
)
plt.show()

X2 = [80, 100, 120, 140, 160, 180, 200]
Y2 = [70, 65, 90, 95, 110, 115, 120]

plt.plot(
    X2
  , Y2
  , color = "green"
  , marker = "o"
  , linestyle = "solid"
)
plt.show()

\smallskip \textbf{GuideLLM:} To create multiple plots in Matplotlib while keeping the limits consistent, follow these steps:

1. Import Libraries: Import matplotlib.pyplot as plt.

2. Plot the First Graph: Use plt.plot() to plot the first data set and display it with plt.show().

3. Adjust Plot Limits: After the first plot, set consistent limits using plt.xlim() and plt.ylim().

4. Plot the Second Graph: Plot the next data set with plt.plot(), reusing the set limits, and display with plt.show().

5. Repeat as Needed: Continue plotting more graphs, maintaining the same axis limits for visual consistency.

This method ensures all plots are coherent and uniformly scaled.

\smallskip
\textbf{Instruction to Model:}  
\textit{Using the provided GuideLLM Response as a reference, generate a solution for the problem outlined above. The solution should implement the outlined steps and methodology clearly, ensuring correctness and adherence to the problem's requirements.
Provide the final implementation or a detailed solution as appropriate.}
\end{minipage}
}
\bigskip

Each input to SolverLLM comprises four components: the problem statement, a natural language task description, the step-by-step methodology generated by GuideLLM, and an instruction prompting code generation aligned with the provided reasoning. This standardized input format reduces ambiguity, promotes adherence to coding best practices, and improves consistency across domains. As illustrated in \textbf{Example 2}, the model is presented with a clearly defined context and is guided to generate a solution that closely follows the outlined methodology. This structured prompting encourages reliable and interpretable outputs while reducing the likelihood of hallucinated or misaligned code. The GuideLLM response effectively serves as a scaffold, supporting SolverLLM in preserving the logical flow and intent of the task.

By integrating structured reasoning with a domain-adapted model, SolverLLM significantly improves the quality, clarity, and reliability of generated code, especially for tasks that require multi-step planning or complex problem decomposition.

\bigskip
\fbox{%
\begin{minipage}{0.9\columnwidth}
\textbf{Example 3: JudgeLLM Reasoning Prompt}

\smallskip
\textbf{Problem Statement:} how to create multiple plots?

\smallskip
\textbf{Description:} I'm to Python and learning it by doing. I want to make two plots with matplotlib in Python. The second plot keeps the limits of first one. Wonder how I can change the limits of each next plot from previous. Any help, please. What is the recommended method? 

X1 = [80, 100, 120, 140, 160, 180, 200, 220, 240, 260]

Y1 = [70, 65, 90, 95, 110, 115, 120, 140, 155, 150]

from matplotlib import pyplot as plt

plt.plot(
    X1
  , Y1
  , color = "green"
  , marker = "o"
  , linestyle = "solid"
)
plt.show()

X2 = [80, 100, 120, 140, 160, 180, 200]
Y2 = [70, 65, 90, 95, 110, 115, 120]

plt.plot(
    X2
  , Y2
  , color = "green"
  , marker = "o"
  , linestyle = "solid"
)
plt.show()

\smallskip \textbf{GuideLLM:} GuideLLM response for specific problem statement.

\smallskip \textbf{SolverLLM:} SolverLLM Answer for specific problem statement.

\smallskip
\textbf{Instruction to Model:}  
\textit{Using the provided Problem Statement, Description, and GuideLLM as references, evaluate the outcomes of the SolverLLM. Specifically assess the solution's correctness, completeness, and adherence to the outlined methodological guidance. Highlight key strengths and clearly identify any weaknesses or areas needing improvement. Provide constructive feedback and assign ratings for accuracy, clarity, and efficiency on a scale of 1 to 5, where 1 is poor and 5 is excellent.}
\end{minipage}
}
\bigskip
\subsection{Automated JudgeLLM Evaluation}\label{judgellm}

The \textbf{JudgeLLM} component is designed to provide rigorous and objective evaluation of solutions generated by the SolverLLM within the RefactorCoderQA benchmark. Powered by GPT-4o, JudgeLLM systematically assesses outputs across multiple dimensions, correctness, clarity, efficiency, and adherence to the structured guidance produced by GuideLLM. GPT-4o was chosen for JudgeLLM due to its strong reasoning ability, reliable code comprehension, and consistent performance in evaluating functional correctness and qualitative improvements. Moreover, JudgeLLM operates using a standardized evaluation prompt and follows a structured, three-stage assessment process:

\begin{itemize} \item \textit{Correctness:} JudgeLLM rigorously verifies whether the provided solution accurately addresses the given problem by checking it against pre-defined test cases, expected outcomes, and domain-specific standards.

\item \textit{Clarity:} JudgeLLM carefully assesses the solution's readability, interpretability, and documentation quality. It evaluates whether the solution's explanations and implementations are logically structured, and easy to follow.

\item \textit{Efficiency:} JudgeLLM analyzes both the solutions' computational and conceptual efficiency, explicitly considering factors such as algorithmic complexity, execution performance, and optimal resource usage.
\end{itemize}

This transparent and structured evaluation methodology enables consistent and meaningful feedback, enhancing both the interpretability of the model’s reasoning and the practical relevance of its solutions. As a result, JudgeLLM significantly strengthens the overall robustness and reliability of the RefactorCoderQA evaluation pipeline.

\section{Experiments}
In this section, we describe the experimental framework used to evaluate the effectiveness and performance of the proposed RefactorCoder-MoE model on the RefactorCoderQA benchmark. The evaluation is conducted systematically across a diverse set of coding tasks spanning four domains: SE, DS, NLP, and ML. Multiple LLMs are tested under a unified setup to ensure fair comparison and comprehensive analysis.

\subsection{Task Formulation}

We formulate \textbf{RefactorCoderQA} as a structured code generation benchmark for training and evaluating LLMs on real-world, domain-specific programming tasks. Each instance in the dataset is represented as a triple:

\begin{equation}
x_i = (q_i, d_i, a_i)
\end{equation}

where $q_i$ is a coding-related problem statement sourced from Stack Overflow, $d_i$ is a detailed task description paired with structured guidance generated by GuideLLM, and $a_i$ is the reference solution derived from the accepted community answer. These triples are used to fine-tune our model, \textbf{RefactorCoder-MoE}, enabling it to align its solution-generation process with expert-driven, community-approved answers.

The objective is to train the model to generate a solution $\hat{a}_i$ conditioned on $(q_i, d_i)$:

\begin{equation}
\hat{a}_i = \mathcal{M}(q_i, d_i),
\end{equation}
where $\mathcal{M}$ denotes the fine-tuned RefactorCoder-MoE model. During inference, model outputs $\hat{a}_i$ are evaluated against reference solutions $a_i$ using both automatic metrics and human judgments. The use of Stack Overflow accepted answers as supervision provides high-quality ground truth and ensures that learning is grounded in real-world developer scenarios.

\subsection{Models}

We evaluate \textit{RefactorCoder-MoE} alongside a diverse set of LLMs, including both closed-source commercial APIs and open-source foundation models. This comprehensive comparison enables performance assessment across varying model sizes, architectures, and access modalities. Specifically, we include GPT-4o and GPT-3.5 Turbo from OpenAI~\cite{achiam2023gpt}, Claude 3.5 Opus from Anthropic~\cite{anthropic2024introducing}, and Gemini 1.5 Pro from Google~\cite{team2024gemini} as high-performing closed-source baselines. On the open-source side, we evaluate models such as DeepSeek-Coder~\cite{guo2024deepseek}, CodeLlama~\cite{roziere2023code}, StarCoder2~\cite{lozhkov2024starcoder}, Mistral~\cite{jiang2023mistral}, Gemma~\cite{team2024gemma}, Phi-2~\cite{abdin2024phi}, Falcon~\cite{penedo2023falcon}, and LLaMA~\cite{touvron2023llama}, covering a range from 1B to 7B parameters and supporting both instruction-following and code generation tasks. For smaller-scale baselines, we include quantized and distilled variants such as DeepSeek-Coder-1.3B and Gemma-2B. All open-source models are evaluated in direct inference mode, while commercial APIs are used in zero-shot settings without additional prompt engineering unless otherwise specified. Our fine-tuned model, \textit{RefactorCoder-MoE}, is trained on the DeepSeek-Coder-7B-Instruct-v1.5 checkpoint and optimized using QLoRA for efficient domain-specific adaptation. This unified evaluation setup ensures fair and consistent comparisons between general-purpose foundation models and domain-specialized fine-tuned models.

All models were evaluated end-to-end on the RefactorCoderQA benchmark under identical input and evaluation protocols. Commercial APIs were evaluated in zero-shot settings, while open-source models were evaluated using direct inference without additional prompt engineering. RefactorCoder-MoE is the only fine-tuned model in this comparison. To ensure fairness and reproducibility, we restrict comparisons to models that can be executed consistently across the full benchmark.

\begin{table}[t!]
\centering
\footnotesize
\caption{Accuracy (\%) of models on the RefactorCoderQA benchmark, computed over the full dataset across all four evaluation domains and overall.}
\label{tab:main-results}
\resizebox{\linewidth}{!}{
\begin{tabular}{l|ccccc}
\toprule
\rowcolor{lightgray}
\textbf{Model} & \textbf{SEQA} & \textbf{NLPQA} & \textbf{MLQA} & \textbf{DSQA} & \textbf{Overall} \\
\midrule
\rowcolor{lightyellow}
\multicolumn{6}{l}{\textit{\textbf{Closed-Source Models}}} \\
GPT-4o Turbo & 73.02 & 63.93 & 80.95 & 74.67 & 70.23 \\
GPT-3.5 Turbo & 57.14 & 59.02 & 65.08 & 50.67 & 53.82 \\
Claude 3.5 Opus & 65.44 & 56.33 & 69.04 & 53.55 & 57.88 \\
Gemini 1.5 Pro & 59.55 & 58.77 & 67.43 & 54.86 & 56.89 \\
\hline
\rowcolor{lightblue}
\multicolumn{6}{l}{\textit{\textbf{Open-Source Models (Medium)}}} \\
DeepSeek-Coder-7B & 31.75 & 24.59 & 28.57 & 41.33 & 30.53 \\
DeepSeek-Coder-6.7B & 28.57 & 36.07 & 25.40 & 32.00 & 24.81 \\
DeepSeek-LLM-7B & 20.60 & 14.75 & 20.63 & 29.33 & 22.14 \\
DeepSeek-R1-Distill-Qwen-7B & 20.63 & 19.67 & 23.81 & 18.67 & 20.99 \\
DeepSeek-R1-Distill-Llama-8B & 19.05 & 24.59 & 20.63 & 24.00 & 23.66 \\
CodeLlama-7B-HF & 22.22 & 19.67 & 26.67 & 20.63 & 20.23 \\
Falcon-7B-Instruct & 17.46 & 13.11 & 19.05 & 26.67 & 21.37 \\
Gemma-7B & 17.46 & 19.67 & 25.40 & 30.67 & 19.08 \\
Phi-2 & 17.46 & 19.67 & 17.46 & 21.33 & 22.90 \\
Mistral-7B & 15.87 & 16.39 & 14.29 & 21.33 & 17.56 \\
StarCoder2-7B & 15.87 & 18.03 & 19.05 & 22.67 & 19.85 \\
\hline
\rowcolor{lightblue}
\multicolumn{6}{l}{\textit{\textbf{Open-Source Models (Small)}}} \\
DeepSeek-Coder-1.3B & 22.22 & 16.39 & 23.81 & 34.67 & 22.52 \\
DeepSeek-R1-Distill-Qwen-1.5B & 6.35 & 14.75 & 15.87 & 14.67 & 13.36 \\
LLaMA-3.2-3B & 14.29 & 14.75 & 12.70 & 24.00 & 16.41 \\
Gemma-2B & 17.46 & 14.75 & 15.87 & 26.67 & 19.85 \\
StarCoder2-3B & 15.87 & 18.03 & 15.87 & 18.67 & 16.41 \\
\hline
\rowcolor{lightgray}
\multicolumn{6}{l}{\textit{\textbf{Fine-Tuned Model}}} \\
\textbf{RefactorCoder-MoE (Ours)} & \textbf{74.68} & \textbf{75.70} & \textbf{83.01} & \textbf{77.32} & \textbf{76.84} \\
\bottomrule
\end{tabular}}
\end{table}

\begin{table*}[t!]
\centering
\small
\caption{Representative Failure Cases Analysis Across Domains.}
\label{tab:failure-cases}
\begin{tabular}{p{0.8cm} p{4cm} p{4.2cm} p{2cm} p{5cm}}
\toprule
\textbf{Domain} & \textbf{Problem Statement} & \textbf{Model Output (Main Phrase)} & \textbf{JudgeLLM Score} & \textbf{Human Comments} \\
\midrule
SE & Debugging a Python segmentation fault & ``Try reinstalling Python or restarting the IDE.'' & Acc: 2/5, Clar: 3/5, Eff: 2/5 & Generic suggestion; missed identifying the memory misuse in the C extension. \\
DS & Pandas groupby with custom aggregation & ``Use \texttt{.groupby().mean()} for the column.'' & Acc: 3/5, Clar: 4/5, Eff: 3/5 & Ignored rows with NaN values; output incorrect for missing data. \\
ML & Hyperparameter tuning in scikit-learn & ``Reduce the learning rate to improve accuracy.'' & Acc: 2/5, Clar: 3/5, Eff: 3/5 & Failed to mention validation folds; oversimplified solution. \\
NLP & Tokenization with Hugging Face & ``Use \texttt{tokenizer.encode()} for preprocessing.'' & Acc: 2/5, Clar: 4/5, Eff: 2/5 & Relied on deprecated API; incompatible with recent library versions. \\
\bottomrule
\end{tabular}
\end{table*}

\subsection{Evaluation Metric}

To evaluate model performance on the RefactorCoderQA benchmark, we adopt a multiple-choice question-answering (MCQA) format. Each test instance comprises a coding-related problem, a set of candidate solutions, and a single ground-truth correct answer. This setup enables consistent and automated evaluation across all models under comparison.

We use \textbf{accuracy} as the primary evaluation metric. For each instance, a model selects one candidate solution, and the prediction is considered correct if it matches the ground-truth label. The overall accuracy is then calculated as the proportion of correctly answered questions out of the total number of test instances:

\begin{equation}
\text{Accuracy} = \frac{1}{N} \sum_{i=1}^{N} \mathbb{1}[\hat{a}_i = a_i]
\end{equation}

where $N$ is the number of test questions, $\hat{a}_i$ is the model’s predicted answer, and $a_i$ is the correct answer. Each candidate's answer is constructed using outputs from multiple models, including RefactorCoder-MoE, GPT-4o, GPT-3.5 Turbo, Claude, Gemini, and others, while the correct option is verified against the accepted answer from Stack Overflow. This setup ensures that model performance is benchmarked not only against human-validated solutions but also in competitive comparison with peer models. We report accuracy at both the domain level, SEQA, DSQA, MLQA, NLPQA, and the aggregate OverallQA level. All models are evaluated in the test set of 261 examples, ensuring a fair and reproducible comparison.

\subsection{Main Results}
Table~\ref{tab:main-results} presents the accuracy of all evaluated models on the RefactorCoderQA benchmark across four technical domains and the overall test set. Our fine-tuned model, \textbf{RefactorCoder-MoE}, achieves the highest performance with an overall accuracy of 76.84\%, significantly outperforming both closed-source and open-source baselines. Notably, it demonstrates strong generalization across all domains, achieving 74.68\% on SEQA, 75.70\% on NLPQA, 83.01\% on MLQA, and 77.32\% on DSQA.

\begin{figure}[t!]
\centering
\includegraphics[width=\linewidth]{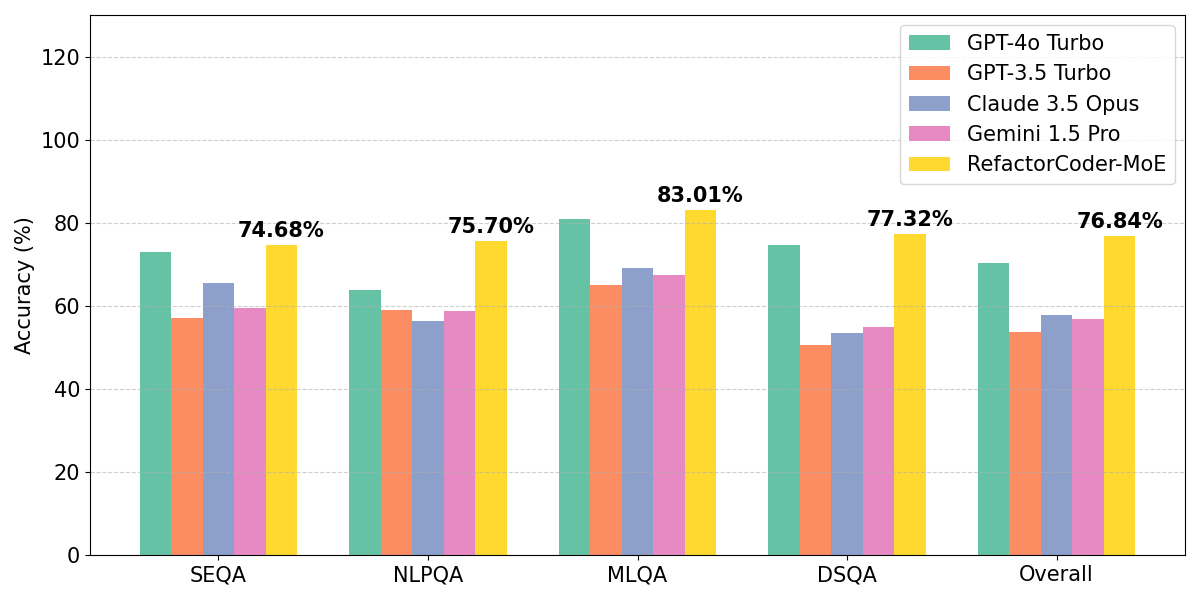}
\caption{
Accuracy Across Domains: Closed-Source Models vs RefactorCoder-MoE
}
\label{fig:clcom}
\end{figure}

Among the closed-source models evaluated, GPT-4o achieves the highest overall accuracy at 70.23\%, followed by Claude 3.5 Opus at 57.88\% and Gemini 1.5 Pro at 56.89\%. GPT-3.5 Turbo ranks lowest among these baselines, with an overall accuracy of 53.82\%. While these models demonstrate reasonably strong performance, they consistently fall short when compared to \textit{RefactorCoder-MoE}, particularly in domains such as MLQA and NLPQA, where the performance gap exceeds five percentage points. \textit{RefactorCoder-MoE} outperforms all closed-source baselines, including GPT-4o, across every evaluated domain. Its advantages are most notable in MLQA and DSQA, where it achieves superior accuracy and demonstrates its strength in handling complex, domain-specific coding tasks, as shown in Fig. \ref{fig:clcom}.

\begin{figure}[t!]
\centering
\includegraphics[width=\linewidth]{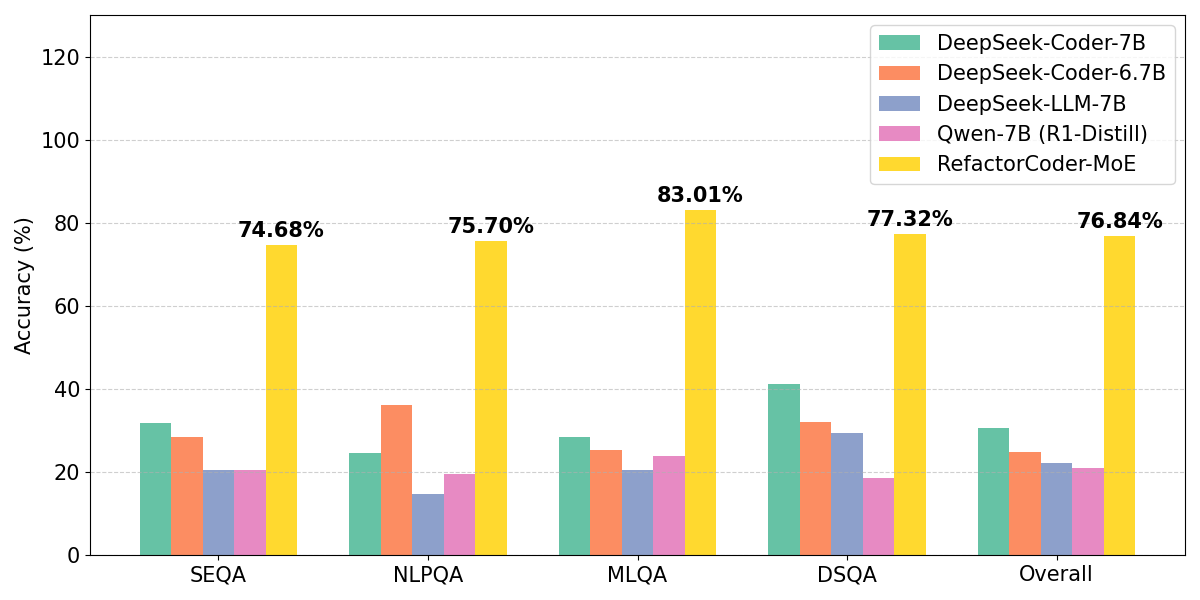}
\caption{
Accuracy Across Domains: Open-Source Models vs RefactorCoder-MoE
}
\label{fig:oscom}
\end{figure}

Open-source models exhibit substantially lower accuracy across all evaluated domains. The strongest open-source baseline, DeepSeek-Coder-7B, achieves an overall accuracy of 30.53\%, followed by DeepSeek-Coder-6.7B at 24.81\% and DeepSeek-LLM-7B at 22.14\%. Smaller models, including Gemma-2B, StarCoder2-3B, and LLaMA-3.2-3B, generally fall below 20\% accuracy, reflecting their limited capacity to solve multi-domain technical QA tasks without domain-specific fine-tuning. In contrast, \textit{RefactorCoder-MoE} delivers significantly higher accuracy across all domains, clearly demonstrating the effectiveness of targeted fine-tuning and structured prompting strategies. These results are illustrated in Fig.~\ref{fig:oscom}.

These results underscore the effectiveness of domain-specific fine-tuning and structured prompting. The strong performance of \textit{RefactorCoder-MoE} highlights the benefits of incorporating task-specific instruction-following through \textit{GuideLLM} and multi-stage evaluation via \textit{JudgeLLM}. The substantial performance gap between our model and all evaluated baselines, particularly on real-world, developer-authored questions, emphasizes the importance of grounded reasoning and methodological alignment in coding-focused LLM applications.

\subsection{Quantitative Analysis and Human Evaluation}

We conducted a comprehensive quantitative evaluation of \textbf{RefactorCoder-MoE} on the RefactorCoderQA benchmark, covering four core domains: Software Engineering (SE), Data Science (DS), Machine Learning (ML), and Natural Language Processing (NLP). This analysis combines large-scale automated evaluation with targeted human assessment to validate the reliability of the proposed evaluation protocol.

\textbf{Evaluation Criteria.}
All model responses were assessed along three dimensions: \textbf{Accuracy} (solution correctness), \textbf{Clarity} (readability and interpretability), and \textbf{Efficiency} (code quality and computational optimality). Each dimension was scored on a 1--5 scale. Automated evaluation was performed using \textit{JudgeLLM}, while human evaluation followed the same rubric to ensure direct comparability.

\textbf{Human Evaluation Protocol.}
To validate the reliability of JudgeLLM, we conducted a human evaluation on a uniformly sampled subset of \textbf{100 responses} generated by RefactorCoder-MoE, consisting of \textbf{25 instances} from each domain (SEQA, DSQA, MLQA, and NLPQA). Three expert annotators with domain-relevant technical backgrounds independently rated each response. Final human scores were obtained by averaging ratings across annotators for each metric.

\textbf{Human vs. JudgeLLM Consistency.}
Table~\ref{tab:human-vs-judge} compares human and JudgeLLM scores on the same set of 100 samples. The results show strong alignment across all domains and metrics, with JudgeLLM consistently reporting slightly higher scores. The average variance between human and automated ratings is below 0.2 per metric, confirming that JudgeLLM provides a reliable approximation of human judgment for large-scale evaluation.

\begin{table}[t]
\centering
\small
\caption{Comparison of Human and JudgeLLM evaluation on 100 RefactorCoder-MoE responses (scores out of 5). Human scores are averaged across three annotators.}
\label{tab:human-vs-judge}
\resizebox{0.8\columnwidth}{!}{%
\begin{tabular}{lccc|ccc}
\toprule
\textbf{Domain} & \multicolumn{3}{c|}{\textbf{Human}} & \multicolumn{3}{c}{\textbf{JudgeLLM}} \\
& Acc. & Clar. & Eff. & Acc. & Clar. & Eff. \\
\midrule
SEQA  & 4.69 & 4.67 & 4.52 & 4.80 & 4.88 & 4.69 \\
NLPQA & 4.72 & 4.66 & 4.67 & 4.87 & 4.91 & 4.83 \\
DSQA  & 4.55 & 4.53 & 4.43 & 4.61 & 4.81 & 4.53 \\
MLQA  & 4.55 & 4.71 & 4.42 & 4.69 & 4.76 & 4.53 \\
\midrule
\textbf{Overall} & \textbf{4.63} & \textbf{4.64} & \textbf{4.51} & \textbf{4.74} & \textbf{4.84} & \textbf{4.64} \\
\bottomrule
\end{tabular}
}
\end{table}

\textbf{Overall Quantitative Results.}
Table~\ref{tab:quant-results} presents a performance comparison across models and domains. RefactorCoder-MoE consistently outperforms GPT-4, DeepSeek-Coder-7B, and CodeLLaMA-13B across all four domains, with particularly strong gains in SE and DS tasks. For RefactorCoder-MoE, we report \textbf{human-annotated} average scores (Accuracy 4.63, Clarity 4.64, Efficiency 4.51), reflecting high real-world solution quality. These results demonstrate that the proposed multi-agent instruction-tuning framework produces not only correct solutions but also clear and efficient code.

\begin{table*}[t]
\centering
\small
\caption{Quantitative Performance Comparison across Domains (scores out of 5). RefactorCoder-MoE scores reflect human evaluation (100 samples); all other models are evaluated using JudgeLLM.}
\label{tab:quant-results}
\resizebox{0.95\textwidth}{!}{%
\begin{tabular}{lccc|ccc|ccc|ccc}
\toprule
\multirow{2}{*}{\textbf{Model}} & \multicolumn{3}{c|}{\textbf{SE}} & \multicolumn{3}{c|}{\textbf{DS}} & \multicolumn{3}{c|}{\textbf{ML}} & \multicolumn{3}{c}{\textbf{NLP}} \\
& Acc. & Clar. & Eff. & Acc. & Clar. & Eff. & Acc. & Clar. & Eff. & Acc. & Clar. & Eff. \\
\midrule
GPT-4             & 4.5 & 4.6 & 4.4 & 4.4 & 4.5 & 4.3 & 4.3 & 4.4 & 4.2 & 4.2 & 4.3 & 4.1 \\
DeepSeek-Coder-7B & 4.1 & 4.0 & 3.9 & 4.0 & 3.9 & 3.8 & 3.8 & 3.9 & 3.7 & 3.7 & 3.8 & 3.6 \\
CodeLLaMA-13B     & 3.9 & 3.8 & 3.7 & 3.7 & 3.6 & 3.5 & 3.5 & 3.6 & 3.4 & 3.4 & 3.5 & 3.3 \\
\midrule
\textbf{RefactorCoder-MoE} 
& \textbf{4.63} & \textbf{4.64} & \textbf{4.51} 
& \textbf{4.63} & \textbf{4.64} & \textbf{4.51} 
& \textbf{4.63} & \textbf{4.64} & \textbf{4.51} 
& \textbf{4.63} & \textbf{4.64} & \textbf{4.51} \\
\bottomrule
\end{tabular}
}
\end{table*}

\subsection{Qualitative Error Analysis}

Although RefactorCoder-MoE achieves strong overall performance, we also examined qualitative failure cases to understand its limitations. Table~\ref{tab:failure-cases} highlights representative examples from each domain. For brevity, we report only the \textit{main phrase} of the model’s answer that illustrates the error, rather than the full response. Alongside each example, we provide JudgeLLM scores (accuracy, clarity, efficiency) and human evaluator comments.

This analysis reveals common error patterns: (1) reliance on overly generic fixes, (2) insufficient handling of edge cases, and (3) use of outdated or deprecated methods. These limitations highlight areas where RefactorCoder-MoE can be further improved by incorporating additional training data and domain-specific error handling strategies.

\subsection{Ablation Studies}

To further investigate the performance of \textbf{RefactorCoder-MoE}, we conducted ablation studies along two independent dimensions: 
\textbf{Problem Type (A)}, which covers 11 categories drawn from RefactorCoderQA (Python, C, Java, JavaScript, NumPy, Pandas, Matplotlib, TensorFlow, PyTorch, Scikit-learn, and LLM), and 
\textbf{Answer Type (B)}, which distinguishes between \textit{Coding} tasks that require executable code and \textit{Non-Coding} tasks that require conceptual reasoning, debugging explanations, or methodological guidance. 

\textbf{Module-Level Ablation Analysis:}
In addition to the above dimensions, we explicitly examine the contribution of individual components within the proposed agentic framework. As reported in Table~\ref{tab:ablation-refactorcoder}, we evaluate three variants of RefactorCoder-MoE: 
(i) \textit{RefactorCoder-MoE (Direct)}, which performs single-stage solution generation without any intermediate reasoning or structured guidance; 
(ii) \textit{RefactorCoder-MoE (w/o GuideLLM)}, which follows the same solver and evaluation pipeline but omits the structured methodological plan generated by GuideLLM; and 
(iii) the full agentic framework integrating GuideLLM, SolverLLM, and JudgeLLM. 
Across all problem categories and answer types, the full agentic framework consistently outperforms both ablated variants. Removing GuideLLM leads to a noticeable degradation in accuracy, while direct inference exhibits the largest performance drop. This pattern confirms that the observed gains are not solely attributable to fine-tuning, but are significantly driven by structured methodological guidance and explicit role separation introduced by GuideLLM within the multi-agent pipeline.

\begin{table}[h!]
\centering
\small
\caption{Latency Analysis of GuideLLM, SolverLLM, and JudgeLLM}
\label{tab:lat-guide-solver-judge}
\resizebox{0.8\columnwidth}{!}{%
\begin{tabular}{lccc}
\toprule
\multirow{2}{*}{\textbf{Domain}} & \multicolumn{3}{c}{\textbf{Inference Latency (in sec)}} \\
\cmidrule(lr){2-4}
                                 & \textbf{GuideLLM} & \textbf{SolverLLM} & \textbf{JudgeLLM} \\
\midrule
SEQA  & 6.425 & 5.919 & 1.987 \\
NLPQA & 6.096 & 6.220  & 2.008 \\
DSQA  & 4.946 & 5.336  & 1.999 \\
MLQA  & 5.758 & 6.015  & 2.824 \\

\bottomrule
\end{tabular}
}
\end{table}

\begin{table*}[t!]
\centering
\small
\caption{Ablation Study Results on RefactorCoderQA across
two independent dimensions: 
\textcolor{black}{\colorbox[HTML]{DAE8FC}{\textbf{Problem Type (A)}}} and 
\textcolor{black}{\colorbox[HTML]{FFF2CC}{\textbf{Answer Type (B)}}}.}
\label{tab:ablation-refactorcoder}
\resizebox{0.95\textwidth}{!}{
\begin{tabular}{
    l|
    >{\columncolor[HTML]{DAE8FC}}c
    >{\columncolor[HTML]{DAE8FC}}c
    >{\columncolor[HTML]{DAE8FC}}c
    >{\columncolor[HTML]{DAE8FC}}c
    >{\columncolor[HTML]{DAE8FC}}c
    >{\columncolor[HTML]{DAE8FC}}c
    >{\columncolor[HTML]{DAE8FC}}c
    >{\columncolor[HTML]{DAE8FC}}c
    >{\columncolor[HTML]{DAE8FC}}c
    >{\columncolor[HTML]{DAE8FC}}c
    >{\columncolor[HTML]{DAE8FC}}c|
    >{\columncolor[HTML]{FFF2CC}}c
    >{\columncolor[HTML]{FFF2CC}}c
}
\toprule
& \multicolumn{11}{c|}{\textbf{Problem Type (A)}} & \multicolumn{2}{c}{\textbf{Answer Type (B)}} \\
\cmidrule(lr){2-12}\cmidrule(lr){13-14}
\textbf{Model} & Python & C & Java & JS & NumPy & Pandas & Matplotlib & TensorFlow & PyTorch & Sklearn & LLM 
& Coding & Non-Coding \\
\midrule
\multicolumn{14}{l}{\textit{Closed-Source Models}} \\
\midrule
GPT-4o Turbo         & 73.5 & 72.4 & 71.2 & 69.8 & 72.1 & 74.0 & 73.2 & 71.9 & 70.8 & 72.4 & 69.9 & 72.8 & 68.5 \\
Claude 3.5 Opus      & 65.0 & 64.3 & 63.8 & 61.2 & 63.5 & 65.4 & 64.7 & 63.1 & 62.8 & 64.5 & 61.0 & 64.0 & 60.2 \\
Gemini 1.5 Pro       & 60.1 & 59.4 & 58.7 & 57.2 & 59.0 & 61.1 & 60.3 & 58.8 & 57.5 & 59.9 & 56.2 & 59.2 & 55.4 \\
GPT-3.5 Turbo        & 57.4 & 56.2 & 55.0 & 54.8 & 55.1 & 56.5 & 55.7 & 54.3 & 53.9 & 55.8 & 52.4 & 56.1 & 51.0 \\
\midrule
\multicolumn{14}{l}{\textit{Open-Source Models}} \\
\midrule
DeepSeek-Coder-7B    & 32.2 & 31.1 & 30.5 & 29.4 & 30.7 & 32.0 & 31.3 & 30.1 & 29.6 & 31.2 & 28.9 & 31.0 & 28.0 \\
DeepSeek-Coder-6.7B  & 28.4 & 27.1 & 26.5 & 25.7 & 26.2 & 27.9 & 26.8 & 25.4 & 24.9 & 26.7 & 24.0 & 26.2 & 23.5 \\
DeepSeek-LLM-7B      & 21.0 & 20.2 & 19.8 & 19.4 & 19.7 & 20.5 & 20.1 & 19.2 & 18.8 & 19.9 & 18.0 & 20.0 & 17.0 \\
CodeLLaMA-7B-HF      & 22.8 & 21.7 & 21.1 & 20.8 & 21.3 & 22.4 & 21.8 & 20.7 & 20.3 & 21.5 & 19.4 & 21.5 & 18.9 \\
StarCoder2-7B        & 16.5 & 15.9 & 15.4 & 14.8 & 15.2 & 16.1 & 15.7 & 14.6 & 14.2 & 15.5 & 13.8 & 15.0 & 13.0 \\
Mistral-7B           & 16.0 & 15.3 & 14.9 & 14.5 & 14.8 & 15.7 & 15.2 & 14.1 & 13.7 & 15.0 & 13.2 & 14.5 & 12.5 \\
\midrule
\textbf{RefactorCoder-MoE (Direct)} 
& \textbf{53.2} & \textbf{52.4} & \textbf{51.9} & \textbf{50.7} & \textbf{52.8} & \textbf{54.1} & \textbf{53.5} 
& \textbf{51.2} & \textbf{50.9} & \textbf{52.0} & \textbf{49.8} 
& \textbf{54.0} & \textbf{50.0} \\
\textbf{RefactorCoder-MoE (w/o GuideLLM)} 
& \textbf{59.2} & \textbf{58.1} & \textbf{57.5} & \textbf{56.8} & \textbf{59.2} & \textbf{60.0} & \textbf{59.1} 
& \textbf{57.2} & \textbf{58.9} & \textbf{55.9} & \textbf{55.2} 
& \textbf{60.0} & \textbf{56.2} \\
\textbf{RefactorCoder-MoE (Ours)} 
& \textbf{75.1} & \textbf{74.0} & \textbf{73.5} & \textbf{72.8} & \textbf{74.2} & \textbf{76.1} & \textbf{75.4} 
& \textbf{73.8} & \textbf{73.0} & \textbf{74.5} & \textbf{72.0} 
& \textbf{75.2} & \textbf{71.8} \\
\bottomrule
\end{tabular}
}
\end{table*}

\textbf{Problem Type (A):} 
As shown in Table~\ref{tab:ablation-refactorcoder}, RefactorCoder-MoE consistently outperforms both open-source and closed-source baselines across all 11 problem categories. The improvements are especially pronounced in Pandas, Python, and TensorFlow, where our structured multi-agent design and fine-tuned methodology provided clearer reasoning paths and more precise solutions. Even in traditionally challenging categories such as C and LLM-related tasks, RefactorCoder-MoE maintained strong performance relative to competitive baselines.

\textbf{Answer Type (B):} 
For Coding tasks, RefactorCoder-MoE achieved significant gains, reflecting its strength in producing executable and syntactically correct code. Non-Coding tasks, while more challenging, also benefited from the GuideLLM-driven decomposition, enabling the model to deliver concise yet detailed explanations. Compared with GPT-4o Turbo and Claude 3.5 Opus, RefactorCoder-MoE demonstrated higher accuracy and clarity across both answer types, showcasing balanced proficiency in both code generation and conceptual reasoning.

Overall, our analysis highlights the versatility of RefactorCoder-MoE. Its ability to handle a wide variety of problem types while excelling in both Coding and Non-Coding tasks underscores the effectiveness of our multi-agent pipeline. By combining domain-specific fine-tuning with structured guidance, RefactorCoder-MoE delivers robust and generalizable improvements across diverse programming challenges.

\subsection{Latency Analysis}
Table \ref{tab:lat-guide-solver-judge} presents the average inference latency (in seconds) across four programming domains for GuideLLM, SolverLLM, and JudgeLLM. These values represent the mean latency per question within each domain. In a conventional setup, a user query is sent directly to a large language model (LLM) for inference. However, in our proposed architecture, the raw query is first processed by GuideLLM, which reformulates it into a more structured and context-rich version. This refined query is then passed to SolverLLM for final code generation. This two-stage interaction results in nearly double the latency compared to a traditional single-model setup. However, this additional processing time leads to significantly higher accuracy and task alignment across diverse programming domains. In future work, we aim to reduce the overall latency by compressing GuideLLM and SolverLLM or replacing them with smaller, optimized models for real-time or edge deployment scenarios.

\section{Limitations}
While we do not claim universal state-of-the-art performance across all existing LLMs, our results demonstrate strong and consistent improvements over all evaluated baselines within our experimental setting. Although \textit{RefactorCoderQA} offers a comprehensive and structured benchmark for evaluating LLMs in coding tasks, several limitations should be noted. The dataset is primarily constructed from Stack Overflow and similar public sources, which may partially overlap with the pretraining data of large language models, potentially introducing data leakage and impacting the assessment of generalization. As a result, the benchmark may underrepresent less common, emerging, or highly specialized coding scenarios that are not frequently discussed in public forums. Additionally, although the dataset spans four key domains SE, DS, ML, and NLP—its overall size remains moderate. This may limit task diversity and reduce the generalization of findings, especially in subdomains with fewer examples.

Another limitation lies in the evaluation framework, which relies on the \textit{JudgeLLM} component powered by GPT-4o. While JudgeLLM demonstrates strong agreement with human assessments, it assumes consistent and accurate performance across all task types. In cases involving unconventional, creative, or domain-specific solutions, automated scoring may produce inaccurate or inconsistent judgments, which can affect fairness and reliability in evaluation outcomes. Finally, the current evaluation focuses on three core dimensions: correctness, clarity, and efficiency. These are essential for assessing the functional quality of generated code. However, other important attributes, such as maintainability, scalability, robustness, and security, are not directly evaluated. This limits the overall depth and completeness of the assessment. 

On the other hand, a practical limitation relates to computational resources. Our experiments were conducted under a fixed GPU configuration, which restricted evaluation to models that could be executed end-to-end across the full benchmark and limited the maximum context length. While this ensures fair and reproducible comparisons, it prevents direct evaluation of very large models and longer context windows. Future work will explore scaling the framework as additional computational resources become available.

Finally, increased inference latency is also the limitation of this work, as each query passes sequentially through both GuideLLM and SolverLLM. This design, while improving accuracy, results in nearly double the response time compared to single-model baselines. Future work will focus on model compression and architectural optimizations to address this latency overhead.

\section{Conclusion}
In this work, we introduced \textit{RefactorCoderQA}, a comprehensive benchmark designed to rigorously evaluate and enhance the capabilities of large language models (LLMs) in solving practical coding tasks. By leveraging real-world questions and accepted solutions sourced from Stack Overflow, we constructed a robust dataset and developed a structured, multi-agent framework comprising \textit{GuideLLM}, \textit{SolverLLM}, and \textit{JudgeLLM}. This pipeline enables systematic methodological guidance, accurate code generation, and rigorous automated evaluation, thereby improving the reliability and interpretability of LLM-generated solutions. Through extensive experimentation, our fine-tuned model, \textit{RefactorCoder-MoE}, demonstrated consistent superiority over state-of-the-art closed-source and open-source models, achieving an overall accuracy of 76.84\% with significant improvements across all evaluated domains. Additionally, a comprehensive human evaluation confirmed the robustness and validity of our \textit{JudgeLLM}-based assessment method, showing high agreement with expert human judgment. This validates \textit{JudgeLLM} as a reliable and scalable tool for automatic solution evaluation. Future work will focus on expanding the benchmark to additional technical domains, enhancing the reasoning capabilities of each agent, and integrating more advanced methodologies to further bridge the gap between artificial and human-level problem-solving in code generation.

\bibliographystyle{IEEEtran}
\bibliography{custom}
\newpage
\appendix
\subsection{Dataset Overview}
The RefactorCoderQA dataset was meticulously curated to reflect realistic, diverse, and domain-relevant programming challenges, facilitating comprehensive evaluation of agentic LLM frameworks. Unlike synthetic benchmarks, our dataset sources real problem–solution pairs directly from Stack Overflow using its official API, with a strict inclusion criterion: only posts with accepted answers were retained. This ensures that each question is not only grounded in authentic developer context but is also verifiably answerable.

The dataset comprises 2,635 QA pairs across 11 fine-grained technical categories, including Python, C, Java, JavaScript, NumPy, Pandas, Matplotlib, TensorFlow, PyTorch, Scikit-learn, and LLM-related queries. Furthermore, each sample is labeled based on its answer type (coding vs. non-coding), allowing task-specific evaluation.

To maintain data quality, we filtered out questions with incomplete, ambiguous, or opinion-based phrasing. Each retained sample was manually validated and restructured into a standardized JSONL schema, comprising a high-level instruction, detailed input (problem description), and output (accepted answer). This format aligns with modern instruction-tuning objectives and supports flexible adaptation for both generative and evaluative agentic workflows.

This dataset serves a dual role: (1) as a supervised fine-tuning source for RefactorCoder-MoE, and (2) as an evaluation benchmark for agent performance across multiple reasoning dimensions (Accuracy, Clarity, Efficiency). Its realistic and multi-domain nature enables robust, interpretable comparison of closed- and open-source models in code understanding and generation tasks.

\subsection{Token-Length Distribution Analysis}

This appendix provides supporting evidence for the token-length analysis summarized in the main text. Specifically, we report detailed statistics of the token-length distributions for both \textit{GuideLLM} and \textit{SolverLLM} outputs across all RefactorCoderQA domains. These statistics are computed from the testing data used in our evaluation and demonstrate that the selected maximum sequence length of 512 tokens sufficiently covers the vast majority of model outputs.

\begin{table}[h]
\centering
\small
\caption{Token-length statistics of GuideLLM and SolverLLM outputs across RefactorCoderQA domains.}
\label{tab:token-length-stats}
\begin{tabular}{lcccccc}
\toprule
\textbf{Domain} & 
\multicolumn{3}{c}{\textbf{GuideLLM}} & 
\multicolumn{3}{c}{\textbf{SolverLLM}} \\
\cmidrule(lr){2-4} \cmidrule(lr){5-7}
 & Mean & 95th \%ile & Max & Mean & 95th \%ile & Max \\
\midrule
SEQA  & 324 & 393 & 484 & 305 & 399 & 527 \\
DSQA  & 248 & 362 & 391 & 252 & 368 & 657 \\
MLQA  & 290 & 374 & 384 & 284 & 352 & 386 \\
NLPQA & 307 & 394 & 426 & 308 & 421 & 468 \\
\midrule
\textbf{Overall} & \textbf{292} & \textbf{386} & \textbf{484} & \textbf{287} & \textbf{399} & \textbf{657} \\
\bottomrule
\end{tabular}
\end{table}

\subsection{RefactorCoder Agentic Framework Models Prompts}

The RefactorCoder agentic framework employs a three-stage prompting design to improve reasoning, code generation, and evaluation quality across diverse programming tasks. This design ensures interpretability, accuracy, and efficiency by explicitly separating problem understanding, solution generation, and evaluation. Table~\ref{tab:agentic-prompts} outlines the standardized prompts used for the three specialized agents: GuideLLM, SolverLLM, and JudgeLLM.

\begin{table*}[t!]
\centering
\small
\caption{Standardized Prompts for RefactorCoder Agentic Framework Models.}
\label{tab:agentic-prompts}
\resizebox{0.95\textwidth}{!}{
\begin{tabular}{l|p{13cm}}
\toprule
\textbf{Agent} & \textbf{Prompt Structure} \\
\midrule
\textbf{GuideLLM} &
\textbf{Objective:} Generate a structured methodology to solve the given coding problem. \newline
\textbf{Problem Statement:} [Insert Stack Overflow Problem Title] \newline
\textbf{Problem Description:} [Insert Stack Overflow Problem Description] \newline
\textbf{Instruction:} 
``Provide a detailed, step-by-step
methodological guide outlining how to approach and
effectively solve the problem described above. Clearly
identify key steps, logical processes, essential consider-
ations, and conceptual frameworks necessary to address
the task comprehensively. Avoid providing the final code
implementation, focusing instead on a structured, clear,
and actionable methodological outline.'' \\
\midrule
\textbf{SolverLLM} &
\textbf{Objective:} Generate executable and optimized code solutions based on GuideLLM’s methodology. \newline
\textbf{Problem Statement:} [Insert Stack Overflow Problem Title] \newline
\textbf{Problem Description:} [Insert Stack Overflow Problem Description] \newline
\textbf{GuideLLM Response:} [Insert GuideLLM Response] \newline
\textbf{Instruction:} 
``Using the provided GuideLLM
Response as a reference, generate a solution for the
problem outlined above. The solution should implement
the outlined steps and methodology clearly, ensuring
correctness and adherence to the problem’s requirements.
Provide the final implementation or a detailed solution as
appropriate.'' \\
\midrule
\textbf{JudgeLLM} &
\textbf{Objective:} Evaluate SolverLLM’s output for correctness, clarity, and efficiency. \newline
\textbf{Problem Statement:} [Insert Stack Overflow Problem Title] \newline
\textbf{Problem Description:} [Insert Stack Overflow Problem Description] \newline
\textbf{GuideLLM Response:} [Insert GuideLLM Response here] \newline
\textbf{SolverLLM Response:} [Insert SolverLLM Response here] \newline
\textbf{Instruction:} 
``Using the provided Problem State-
ment, Description, and GuideLLM as references, evaluate
the outcomes of the SolverLLM. Specifically assess the
solution’s correctness, completeness, and adherence to the
outlined methodological guidance. Highlight key strengths
and clearly identify any weaknesses or areas needing
improvement. Provide constructive feedback and assign
ratings for accuracy, clarity, and efficiency on a scale of
1 to 5, where 1 is poor and 5 is excellent.'' \\
\bottomrule
\end{tabular}
}
\end{table*}

This structured prompting pipeline enhances problem decomposition, ensures faithful solution generation, and provides transparent evaluation. GuideLLM improves reasoning interpretability, SolverLLM guarantees executable outputs, and JudgeLLM offers consistent scoring comparable to human evaluators.

\subsection{Agentic Models Outcomes}

To illustrate the effectiveness of the RefactorCoder agentic framework, Table~\ref{tab:agentic-outcomes} presents a representative example outcome, showing the structured response from GuideLLM, the executable code from SolverLLM, and the evaluation feedback from JudgeLLM.

\begin{table*}[t!]
\centering
\small
\caption{Representative Outcome of Problem Statement 1 for RefactorCoder Agentic Framework Models}
\label{tab:agentic-outcomes}
\resizebox{0.95\textwidth}{!}{
\begin{tabular}{l|p{12cm}}
\toprule
\textbf{Problem Statement 1} & \textbf{How to create multiple plots?} \\
\midrule
\textbf{Problem Description} & 
I'm new to Python and learning it by doing. I want to make two plots with matplotlib in Python. 
The second plot keeps the limits of the first one. Wonder how I can change the limits of each next plot 
from previous. Any help, please. What is the recommended method? \par
\parbox{\linewidth}{\ttfamily
X1 = [80, 100, 120, 140, 160, 180, 200, 220, 240, 260]\par
Y1 = [70, 65, 90, 95, 110, 115, 120, 140, 155, 150]\par
from matplotlib import pyplot as plt\par
plt.plot(X1, Y1, color="green", marker="o", linestyle="solid")\par
plt.show()\par
X2 = [80, 100, 120, 140, 160, 180, 200]\par
Y2 = [70, 65, 90, 95, 110, 115, 120]\par
plt.plot(X2, Y2, color="green", marker="o", linestyle="solid")\par
plt.show()
} \\
\midrule
\textbf{GuideLLM} & 
To create multiple plots in matplotlib in Python while keeping the limits of the previous plot for the next one, you can follow these steps:\par
1. \textbf{Import Necessary Libraries}: Start by importing the required libraries. In this case, you need to import matplotlib.pyplot as plt.\par
2. \textbf{Plot the First Graph}: Plot the first set of data points using the plt.plot() function. Customize the plot with color, marker, and linestyle options if needed. Display the plot using plt.show().\par
3. \textbf{Adjusting Plot Limits}: After displaying the first plot, you can adjust the limits for the next plot by using plt.xlim() and plt.ylim() functions. Set the limits based on the data points of the first plot.\par
4. \textbf{Plot the Second Graph}: Define the second set of data points and plot them using plt.plot(). Customize the plot similar to the first one and display it with plt.show().\par
5. \textbf{Repeat as Needed}: You can create additional plots following the same approach of adjusting the plot limits after each plot to maintain consistency. \\
\midrule
\textbf{SolverLLM} & 
To create multiple plots in matplotlib in Python while adjusting the limits of each subsequent plot based on the previous one, you can follow the steps outlined below. The solution ensures that the limits of the second plot are kept within the range of the first plot:\par
\parbox{\linewidth}{\ttfamily
from matplotlib import pyplot as plt\par
X1 = [80, 100, 120, 140, 160, 180, 200, 220, 240, 260]\par
Y1 = [70, 65, 90, 95, 110, 115, 120, 140, 155, 150]\par
plt.plot(X1, Y1, color="green", marker="o", linestyle="solid")\par
plt.show()\par
X2 = [80, 100, 120, 140, 160, 180, 200]\par
Y2 = [70, 65, 90, 95, 110, 115, 120]\par
plt.xlim(min(X1), max(X1))\par
plt.ylim(min(Y1), max(Y1))\par
plt.plot(X2, Y2, color="green", marker="o", linestyle="solid")\par
plt.show()
} \\
\midrule
\textbf{JudgeLLM} & 
The solution provided is highly accurate, clearly following the step-by-step plan from GuideLLM. 
The use of \texttt{plt.xlim()} and \texttt{plt.ylim()} correctly ensures that the limits of the second plot are consistent with the first, addressing the problem statement precisely. 
The code is syntactically correct and executable without modification. 
Clarity is excellent, with well-structured code and clear logic. 
Efficiency is strong, as the solution avoids unnecessary operations and uses concise matplotlib functions.\par
\textbf{Scores:} Accuracy = 5, Clarity = 5, Efficiency = 5 \\
\bottomrule
\end{tabular}
}
\end{table*}

\newpage

\end{document}